%% file: main.tex
\documentclass[conference]{IEEEtran}
\usepackage{times}

\usepackage[sorting=none, maxcitenames=1, maxbibnames=8, style=numeric-comp]{biblatex}
\usepackage{multicol}
\usepackage[bookmarks=true]{hyperref}
\usepackage{algorithm}
\usepackage[noend]{algpseudocode}
\usepackage{bm}
\usepackage{eqnarray}

\input{packages.tex}
\title{Train Robots in a JIF: \underline{J}oint \underline{I}nverse and \underline{F}orward Dynamics with Human and Robot Demonstrations}

\author{\authorblockN{Gagan Khandate\authorrefmark{1}\authorrefmark{2},
Boxuan Wang\authorrefmark{1}\authorrefmark{3},
Sarah Park\authorrefmark{1}\authorrefmark{2},
Weizhe Ni\authorrefmark{2}\authorrefmark{3}, \\
Joaquin Palacios\authorrefmark{3}, 
Kathryn Lampo \authorrefmark{3},
Philippe Wu\authorrefmark{3},
Rosh Ho\authorrefmark{3},
Eric Chang\authorrefmark{3} and
Matei Ciocarlie\authorrefmark{3}}
\authorblockA{ \authorrefmark{2}Dept. of Computer Science~~~\authorrefmark{3}Dept. of Mechanical Engineering~~~\authorrefmark{1}joint first authorship\\
Columbia University, New York, NY 10027, USA
\\
Corresponding email:~\texttt{gagank@cs.columbia.edu}
}}
\begin{document}
\makeatletter
\let\@oldmaketitle\@maketitle
\renewcommand{\@maketitle}{\@oldmaketitle
  \includegraphics[width=\linewidth] {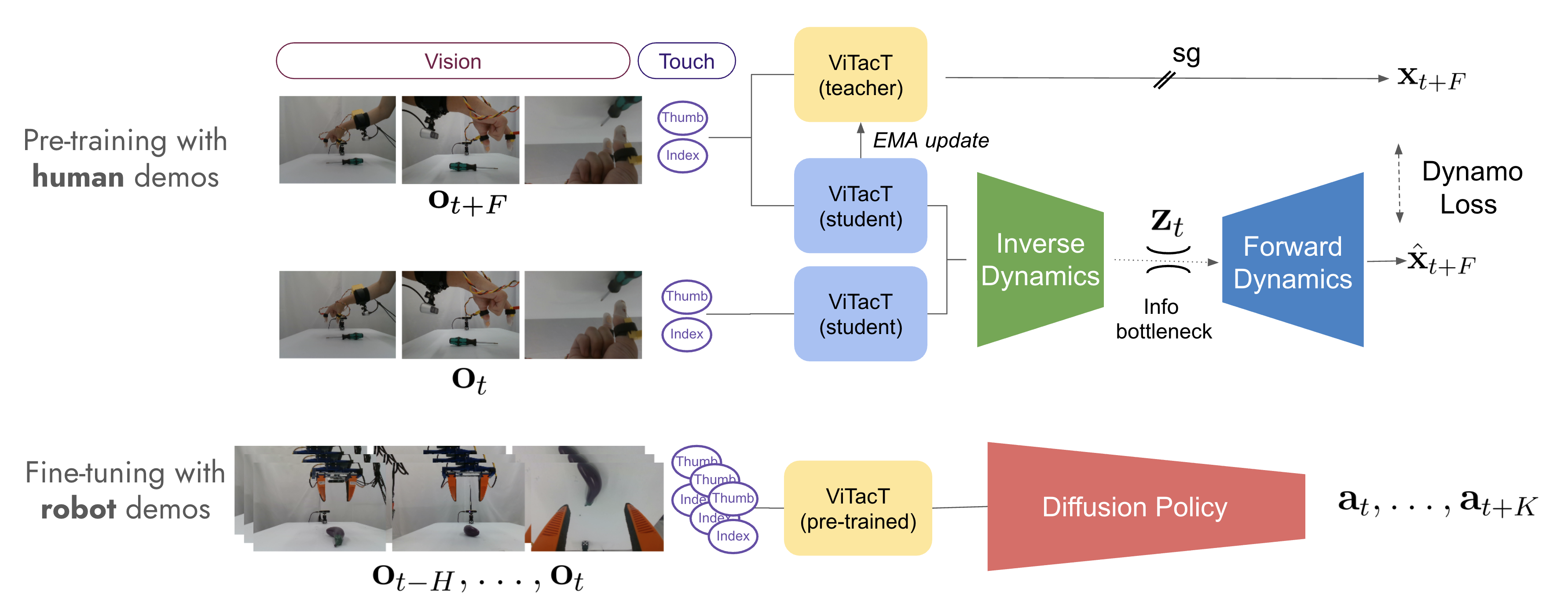} \\[0.25em]
  \refstepcounter{figure}\textbf{Fig. 1:} We introduce a framework for pre-training using multi-modal human demonstrations. We extract latent state representations by \textbf{J}ointly learning \textbf{I}nverse and \textbf{F}orward dynamics - JIF. Dynamics driven state representations maximize manipulation centric information in the human demonstrations allowing efficient and generalizable imitation learning by fine-tuning with a small number of robot demonstrations.}
  \label{fig:real} \medskip \vspace{-10pt}
\makeatother

\maketitle



%


\begin{abstract}
\input{abstract.tex}
\end{abstract}

\IEEEpeerreviewmaketitle

\input{intro.tex}
\input{related.tex}

\input{method.tex}

\input{results.tex}

\printbibliography
\end{document}

%% file: packages.tex
\usepackage{graphicx}
\usepackage[font=small]{caption}
\usepackage{float}
\usepackage{amsmath}
\usepackage{amssymb}
\usepackage[dvipsnames]{xcolor}
\hypersetup{colorlinks,linkcolor={black},citecolor={magenta},urlcolor={magenta}}

\usepackage{algorithm} 
\usepackage{algpseudocode} 
\usepackage{lipsum}
\usepackage{subcaption}

\usepackage{booktabs}

\usepackage{caption}

\newcommand{\futlen}{F}
\newcommand{\histlen}{H}

%% file: abstract.tex
Pre-training on large datasets of robot demonstrations is a powerful technique for learning diverse manipulation skills but is often limited by the high cost and complexity of collecting robot-centric data, especially for tasks requiring tactile feedback. This work addresses these challenges by introducing a novel method for pre-training with multi-modal human demonstrations. Our approach jointly learns inverse and forward dynamics to extract latent state representations, towards learning manipulation specific representations. This enables efficient fine-tuning with only a small number of robot demonstrations, significantly improving data efficiency. Furthermore, our method allows for the use of multi-modal data, such as combination of vision and touch for manipulation. By leveraging latent dynamics modeling and tactile sensing, this approach paves the way for scalable robot manipulation learning based on human demonstrations.

%% file: intro.tex
\section{Introduction}

Pre-training on large datasets of robot demonstrations~\cite{Walke2023-mf, Khazatsky2024-wv} using imitation learning \cite{Chi2023-mz, Ha2023-bl, Fu2024-xt, } or offline reinforcement learning \cite{Kumar2022-rs, Kumar2022-fn} is rapidly becoming a standard technique for learning diverse robot manipulation skills, even pushing towards the development of foundation models \cite{Black2024-uk, Open_X-Embodiment_Collaboration2023-lp, Kim2024-wz, Octo-Model-Team2024-ix, Brohan2023-ol, Shang2024-ba, Mandi2022-oo}. However, these methods rely on extensive collections of robot demonstrations, typically gathered through teleoperation, where the demonstrator remotely operates the robot hardware that is performing the task. This process is expensive to scale; it is also impractical if complex hardware is needed, as is the case for tasks like dexterous manipulation which benefit from real-time tactile feedback for the teleoperator.

In contrast, human demonstrations, where the demonstrator directly performs the task of interest, are significantly less costly to acquire. Furthermore, flexible tactile sensors are making continuous strides~\cite{Luo2021-rk}, and instrumenting human fingertips with such sensors can preserve tactile sensation for the demonstrator while still recording tactile feedback throughout the demonstration, alongside commonly used visual observations. This suggests that pre-training with such multi-modal human demonstrations offers a promising avenue for overcoming the limitations of robot-centric data collection.

Prior visual representation learning work\cite{ Dasari2023-rq, , Pari2021-ur, Majumdar2023-pc, Karamcheti2023-mb, Liu2022-jr, Wang2024-gw, Chen2024-rg, Khandelwal2022-cn}, such as masked visual pre-training (MVP) \cite{Liu2022-jr, Radosavovic2023-hg, Radosavovic2023-fu, Parisi2022-ea}, has explored leveraging visual human demonstrations such as videos for state representation learning. However, these approaches are inherently dynamics-agnostic, and often rely on reconstruction-based objectives that do not capture the underlying system dynamics critical for manipulation tasks. Moreover, the reliance on reconstructing high-dimensional sensory inputs makes these methods computationally expensive and challenging to scale. In contrast, dynamics-driven representation learning provides a more promising alternative by encoding task-relevant information. 

Dynamics-driven representation learning, which involves learning a forward dynamics model, presents challenges when applied to multi-modal human demonstrations. Obtaining action labels, such as hand joint motion, requires costly sensing systems, and existing methods often rely on reconstructing high-dimensional observations, making them computationally expensive. Approaches like LAPO \cite{Schmidt2023-me} mitigate these issues by jointly learning inverse and forward dynamics to capture manipulation-specific information without action labels. Building on this, DynaMo \cite{Cui2024-jf} enhances efficiency through teacher-student distillation while preserving the benefits of dynamics-driven learning. However, these methods focus on using robot demonstrations alone.

We extend this family of approaches to pre-training with multi-modal human demonstrations, enabling representation learning without action labels or costly reconstructions. By adapting the joint inverse-forward dynamics paradigm and incorporating teacher-student distillation, our method significantly improves computational efficiency. This makes representation learning scalable and practical for real-world robotics applications. 

We demonstrate the effectiveness of our representation learning method through imitation learning. Our results show significant improvements in demonstration complexity and generalization, achieving strong performance with a small number of robot demonstrations—tasks that previously required larger datasets. Furthermore, our approach enables the introduction of a novel data collection paradigm: instrumented visuo-tactile human demonstrations. By equipping human demonstrators with tactile sensors, we capture rich tactile information, leading to improved robustness in the learned manipulation skills.

Overall, our method offers a promising path towards scaling robot learning via pre-training with human demonstrations to a wider range of complex tasks, such as multi-fingered manipulation, where traditional teleoperation is difficult. The key contributions of this work are:
\begin{enumerate}
\item We introduce a pre-training framework centered on multi-modal human demonstrations via learning inverse and forward dynamics through teacher-student distillation for computational efficiency.
\item We present a novel paradigm of robot learning from instrumented visuo-tactile human demonstrations, capturing rich tactile information alongside visual data.
\item To the best of our knowledge, we are the first to demonstrate the benefit of incorporating tactile observations in human demonstrations for improving the robustness of learned robot manipulation skills.
\end{enumerate}

%% file: related.tex
\section{Related Work}

Using human demonstrations is a well-established approach in robot learning, often involving video demonstrations to learn reward models for reinforcement learning. For example, 	\textcite{Shao2021-pm} classify human demonstrations into task categories and use the classification score as a reward signal to train a robot policy. Similarly, \textcite{Chen2021-zp} train a discriminator to determine whether two videos depict the same task. By leveraging the similarity score as a reward and combining it with an action-conditioned video prediction model, these methods enable robots to execute new tasks based on human demonstrations. Additionally, ViPER \cite{Escontrela2023-mm} employs a video prediction transformer and uses the log-likelihood of predicted frames as a reward function to guide task execution.

Other works focus on imitation learning, developing frameworks to acquire semantic skills that can transfer across domains \cite{Pertsch2022-yp, Xu2023-jg, Zakka2021-in, Xu2023-dd}. To leverage large-scale human video datasets, such as SomethingSomething-V2 \cite{Goyal2017-xg} and Ego4D \cite{Grauman2021-vl}, self-supervised learning methods have been applied to develop visual representations, demonstrating benefits for downstream policy learning \cite{Dasari2023-rq, Pari2021-ur, Majumdar2023-pc, Karamcheti2023-mb, Liu2022-jr, Wang2024-gw, Chen2024-rg, Khandelwal2022-cn}. For example, 	extcite{Parisi2022-ea} use MoCo representations, while R3M \cite{Nair2022-ss} applies masked autoencoders (MAE). Other works \cite{Liu2022-jr, Radosavovic2023-hg, Radosavovic2023-fu, Parisi2022-ea} explore masked visual pre-training (MVP), though primarily with robot demonstrations. However, reconstruction-based methods like these are computationally expensive and face scalability challenges.

Multi-modal demonstrations are particularly valuable for complex tasks, such as multi-fingered dexterous manipulation, where precise, contact-rich interactions are required. Visual demonstrations alone are often insufficient to capture the detailed sensory feedback needed for these tasks. However, reconstruction based representation learning is too expensive and challenging. Recent methods have sought to overcome these challenges through knowledge self-distillation, a more efficient alternative to reconstruction. Techniques such as BYOL \cite{Grill2020-gr} and DINO \cite{Caron2021-va} employ a teacher-student framework with identical networks, learning invariant representations by aligning predictions across augmented views of the same input. This approach reduces computational overhead, making it particularly suitable for multi-modal data.

Dynamics-driven representation learning has also been explored in various works. Early approaches, such as temporal contrastive learning \cite{Sermanet2017-xs}, learn representations by contrasting temporally adjacent frames. Other methods, such as those proposed by 	\textcite{EdwardsUnknown-lz}, simultaneously learn a world model and a latent policy using visual demonstrations. Similarly, 	\textcite{Schmidt2023-me} enhance representation learning by combining a forward dynamics model with an inverse dynamics model and an information bottleneck. Recently, dynamics-driven representations have been integrated with knowledge self-distillation (DynaMo \cite{Cui2024-jf}) , making these approaches particularly useful when explicit action information is unavailable, as with human demonstrations.

Building on this line of research, our work extends DynaMo \cite{Cui2024-jf} by incorporating multi-modal human demonstrations, including tactile sensing, to improve downstream imitation learning with limited robot demonstrations. A key distinction of our approach is that we are the first to apply dynamics-driven representation learning for multi-modal data and the first to implement self-distillation for such data. Unlike prior efforts that rely on in-the-wild, uni-modal video datasets, we collect task-specific human demonstrations tailored to our downstream tasks. Additionally, we integrate low-cost tactile sensors for instrumented human demonstrations, learning visuo-tactile representations for contact rich tasks.

Although several other works have collected tactile data from human demonstrators \cite{Ruppel2024-ew, Sundaram2019-ju, Sundaram2019-qn, Wei2023-rc, Sagisaka2012-es, Martin2004-dl, Sagisaka2011-tu, Yeo2016-mn, Kim2010-fz, McCaw2018-zv}, these methods have not been used to extract complex robot skills. In contrast, we use low-cost tactile sensors to capture rich multi-modal data and enable imitation learning with a small number of robot demonstrations. Furthermore, we are the first to demonstrate the benefit of incorporating tactile observations in human demonstrations for improving the robustness of learned robot manipulation skills.

Overall, this work explores multi-modal human demonstrations as a foundation for scalable and efficient pre-training, focusing on improving performance in complex, contact-rich tasks for future imitation learning scenarios.

%% file: method.tex
\section{Methods}

In this section, we present our framework for imitation learning, which leverages pre-training with multi-modal human demonstrations to improve the efficiency and effectiveness of policy learning with limited robot demonstrations. Our approach aims to address the challenges associated with learning complex manipulation skills by utilizing a two-stage process: (1) pre-training, where a multi-modal encoder is trained to learn latent state representations from human demonstrations providing multi-modal sensing data but no action labels, and (2) fine-tuning, where a diffusion-based policy is trained using a smaller set of robot demonstrations.

During pre-training, we extract a structured latent state-space by jointly learning forward and inverse dynamics models in the latent space. This allows us to capture task-relevant features from multi-modal human demonstrations without requiring explicit action labels, making the process more scalable and data-efficient. To enhance the quality and robustness of the learned representations, we employ knowledge self-distillation with a teacher-student setup and utilize DynaMo loss to ensure meaningful latent representations.

In the fine-tuning phase, we train a Diffusion Policy conditioned on a history of learned latent state representations alongside a conditioning variable such as a goal or task label to generate actions. By leveraging the pre-trained encoder, our method enables efficient imitation learning, reducing the number of robot demonstrations required to achieve high task success rates.

Our framework effectively combines visual and tactile modalities through our ViTacT multi-modal encoder, which processes multiple camera views and tactile data to produce rich latent representations. In this study, we use convolutional architectures due to their efficiency on smaller-scale datasets, but we anticipate that transformer-based approaches will improve scalability for future applications on larger datasets.

The following subsections provide a detailed explanation of the pre-training and fine-tuning stages, describing the model architectures, loss functions, and training procedures employed in our approach.

\subsection{Problem Definition}

Let $\mathcal{D}_{h}$ be the dataset of observation-only multi-modal human demonstrations consisting of observations $(\mathbf{o}^h_0,\ldots,\mathbf{o}^h_N)$, and let $\mathcal{D}_{r}$ be the data set of action-labeled demonstrations $(\mathbf{o}^r_0,\mathbf{a}^r_0\ldots,\mathbf{a}^r_{N-1}, \mathbf{o}^h_N)$ obtained by teleoperation. Note that the human demonstrations are action free. Given the ease of obtaining human demonstrations and the difficulty of robot demonstrations, we assume $\mathcal{D}_h >> \mathcal{D}_r$. 

Our objective is to efficiently learn a policy for skills demonstrated in $\mathcal{D}_r$ by leveraging pre-training on human demonstrations in $\mathcal{D}_h$. We define the policy as $\pi(\bm{a}_t, \ldots,\bm{a}_{t+K} | \bm{o}_{t}, \ldots,\bm{o}_{t-H}, u)$, where $\bm{o}_{t}, \ldots,\bm{o}_{t-H}$ represent a history of observations, and $u$ is a conditioning variable such as a goal, task label, or language instruction embedding. For simplicity, we omit the embodiment superscripts $h$ and $r$ throughout the discussion.

In the pre-training stage, we seek to pre-train multi-modal encoder $\phi$ by learning dynamics in the latent space. Particularly,  a latent inverse dynamics model $h(\mathbf{z}_t | \mathbf{x}_t, \mathbf{x}_{t+\futlen})$ and a forward dynamics model $f(\mathbf{x}_t | \mathbf{x}_t, \mathbf{z}_{t})$ where $\mathbf{x}_t$ is the latent state of observation $\mathbf{o}_t$ and $\mathbf{z}_t$ is the latent action between the observation pair $\mathbf{x}_t$ and $\mathbf{x}_{t+\futlen}$. Later in the imitation learning stage, we seek to learn a diffusion policy $\psi(\bm{a}_t, \ldots,\bm{a}_{t+K} | \bm{x}_{t}, \ldots,\bm{x}_{t-\histlen}, u)$ that generates the actions given a history of observations.

\subsection{Pre-training with Multi-modal Human Demonstrations}
\label{sec:stage1}


In the pre-training phase, we aim to model dynamics in the latent space to avoid a computationally expensive reconstruction. Let $\bm{x}_t$ and $\bm{x}_{t+\futlen}$ denote the latent representations of observations $\bm{o}_t$ and $\bm{o}_{t+\futlen}$, respectively, obtained through a multi-modal encoder $\phi$. 

Specifically, we learn a forward dynamics model \( f(\hat{\bm{x}}_{t+\futlen} | \bm{x}_t, \bm{z}_t) \) while simultaneously, we train an inverse dynamics model \( h(\bm{z}_t | \bm{x}_t, \bm{x}_{t+\futlen}) \). The training objective minimizes the prediction error in the latent space:

\[
d(\bm{x}_{t+\futlen}, \bm{\hat{x}}_{t+\futlen}),
\]
where \( d \) is a similarity measure and and $F=6$. The low-dimensional latent action \(\bm{z}_t\), introduced by the inverse dynamics model, serves as an information bottleneck, which is essential for efficient representation learning \cite{Schmidt2023-me}. However, naively minimizing the latent state prediction loss,
\[
-\mathbb{E}[d(\bm{x}_{t+\futlen}, \bm{\hat{x}}_{t+\futlen})],
\]
can lead to latent collapse.

To address this, we employ knowledge self-distillation, learning two identical encoders: a teacher network, \(\phi_{teacher}\), and a student network, \(\phi_{student}\). This approach has been shown to improve representation quality while reducing memory requirements and training time, especially with high-dimensional observations. The latent state \(\bm{x}_{t+\futlen}\) is computed using the teacher network:
\[
\bm{x}_{t+\futlen} = \phi_{teacher}(\bm{o}_{t+\futlen}).
\]

The predicted latent state \(\hat{\bm{x}}_{t+\futlen}\) is computed as follows:
\begin{enumerate}
    \item Compute the student embedding: \(\phi_{student}(\bm{o}_t)\).
    \item Predict the forward dynamics:
    \[
    \hat{\bm{x}}_{t+\futlen} \sim f(\hat{\bm{x}}_{t+\futlen} | \bm{x}_t, \bm{z}_t),
    \]
    where \(\bm{z}_t\) is sampled from the inverse dynamics model:
    \[
    \bm{z}_t \sim h(\bm{z}_t | \bm{x}_t, \bm{x}_{t+\futlen}).
    \]
\end{enumerate}

To ensure stability and effective learning, we minimize the latent state prediction loss alongside regularization components:
\[
-\mathbb{E}[d(\text{sg}[\bm{x}_{t+\futlen}], \bm{\hat{x}}_{t+\futlen})] + \beta d_{KL}(\mathbf{z}_t || \mathcal{N}(0,1))
\]
where \(\text{sg}\) indicates a stop-gradient operation to prevent teacher updates from the gradient of this loss and $d_{KL}$ is the KL-divergence. Finally, the teacher encoder \(\phi_{teacher}\) is updated as an Exponential Moving Average (EMA) of the student encoder \(\phi_{student}\). This enables knowledge self-distillation for learning state-representations.

While various similarity loss can be used, we use Dynamo Loss as it has been successfully used for dynamics driven representation learning. We therefore minimize \[
\mathcal{L}_{Dynamo} + \beta d_{KL}(\mathbf{z}_t || \mathcal{N}(0,1))
.\]
Dynamo Loss  comprises two components: cosine similarity  and covariance regularization terms. The cosine similarity is defined as \[
\mathcal{L}_{\text{cosine}}(\hat{\mathbf{x}}_{t+F}, \mathbf{x}_{t+F}) = 1 - \frac{\langle \hat{\mathbf{x}}_{t+F}, \text{sg}[\mathbf{x}_{t+F}] \rangle}{\|\hat{\mathbf{x}}_{t+F}\|_2 \cdot \|\text{sg}[\mathbf{x}_{t+F}]\|_2}.
\] 
This loss penalizes when the two vectors deviate from each other. The second loss component is the covariance regularization loss defined as \[
\mathcal{L}_{\text{cov}}(\hat{\mathbf{X}}_{t+F}) = \frac{1}{d} \sum_{i \neq j} \left[\text{Cov}(\hat{\mathbf{X}}_{t+F})\right]_{i,j}^2,
\]
where \(\text{Cov}(\hat{\mathbf{X}}_{t+F})\) is the covariance matrix of predicted latent states \(\hat{\mathbf{X}}_{t+F}\), and \(d\) is the dimensionality of the feature representations. This term minimizes the off-diagonal elements of the covariance matrix, thereby encouraging feature decorrelation and reducing redundancy among the learned features. Therefore, the total DynaMo loss is the weighted sum of the two components \[
\mathcal{L}_{Dynamo} = \mathcal{L}_{\text{cosine}} + \lambda \mathcal{L}_{\text{cov}},
\] where \(\lambda\) is a hyperparamter to control the tradeoff between the two components. The default value of \(\lambda\) is 0.04.

In our method, $\phi$ is a ViTacT multi-modal encoder, which processes multiple camera views and tactile modalities to generate latent representations. Each camera view is encoded into a token using a ResNet-based embedding, while tactile signals are processed through 1D convolution to produce tactile embeddings. These embeddings are then used as inputs to a transformer decoder, enabling the fusion of multi-modal information for downstream policy learning. However, in large-scale applications with more extensive training data, we anticipate that patch-based encoding, as implemented in Vision Transformers (ViTs), will offer superior performance due to the favorable scaling properties of transformer architectures compared to convolutional networks. In our approach, we utilize CNN-based feature extraction, as our work focuses on a relatively smaller scale of data, where convolutional architectures remain effective and computationally efficient.

\subsection{Imitation Learning with Robot Demonstrations}


In the second imitation learning phase, we train a diffusion policy conditioned on the latent state representations extracted by the pre-trained multi-modal encoder $\phi$, facilitating efficient learning from a limited number of demonstrations. Specifically, the diffusion policy $\psi(\bm{a}_t, \ldots, \bm{a}_{t+K} \mid \bm{x}_t, \ldots, \bm{x}_{t-\histlen}, u)$ generates actions based on a history of latent states and a conditioning variable $u$, such as a goal, task label, or language instruction. We use a history $H=16$ and action chunk length $K=16$ in our experiments. 

By leveraging the pre-trained encoder $\phi$, our approach effectively utilizes human demonstration data to provide informative latent representations, improving the efficiency of imitation learning. As shown in our results, this leads to faster convergence and better generalization with fewer robot demonstrations.

%% file: results.tex
\section{Experimental Results}

\begin{figure}
    \centering
    \includegraphics[trim=18mm 3mm 0mm 0mm,clip=true,width=0.99\linewidth]{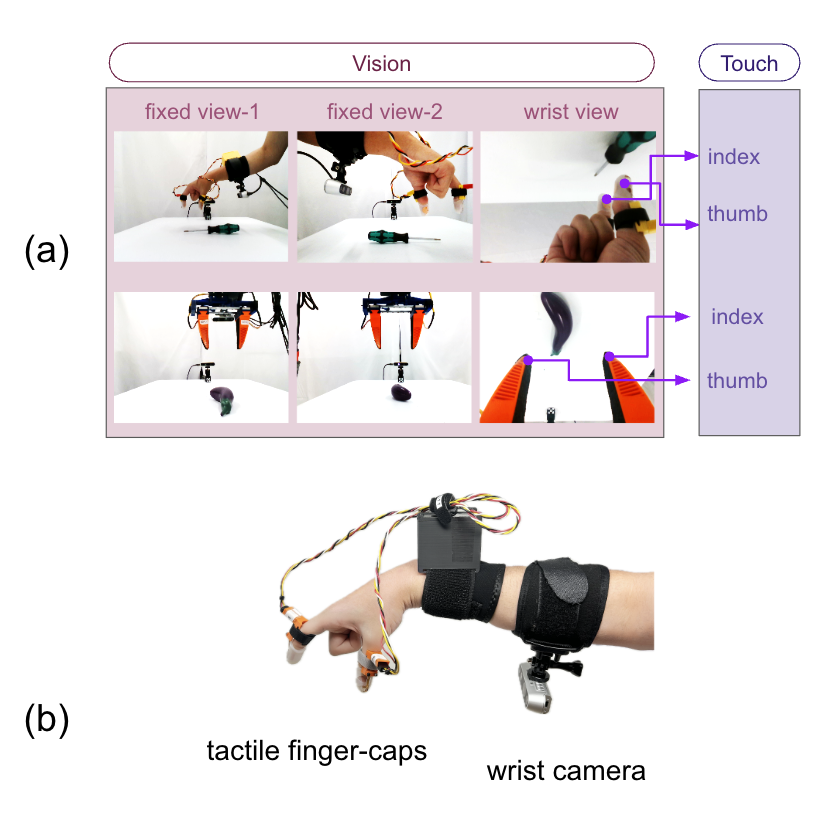}
    \caption{Human and Robot Data Collection Setup. The robot setup includes three camera views—two side views and one wrist view—and a two-fingered gripper with embedded tactile sensors at the fingertips. The human data collection setup mirrors the robot's camera configuration, with tactile data collected using a fingertip cap device equipped with a Singletact capacitive sensor on index and the thumb fingers.}
    \label{fig:demosetup}
\end{figure}

\begin{figure}
    \centering
    \includegraphics[width=\linewidth]{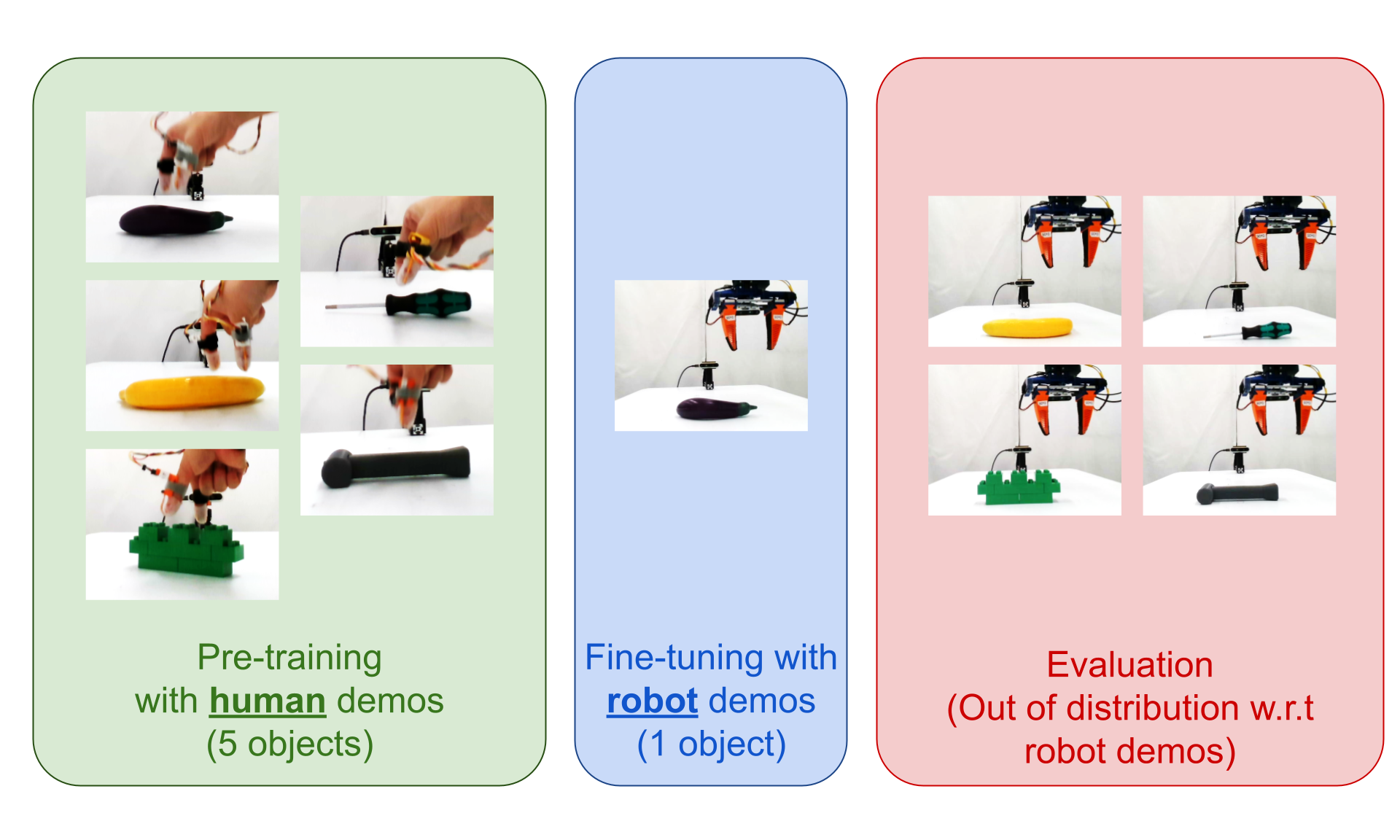}
    \caption{Objects used for pre-training, fine-tuning, and generalization evaluation. As illustrated, we pre-train using human demonstrations collected for all five objects but fine-tune the imitation learning model using only one object. To evaluate generalization, we test on the remaining four objects, making the evaluation out-of-distribution relative to the robot fine-tuning phase but in-distribution with respect to the human demonstrations used during pre-training.}
    \label{fig:demo-objects}
\end{figure}

In this section, we evaluate the effectiveness of our proposed pre-training method using human demonstrations, focusing on its data efficiency and robustness. Additionally, we assess the advantages of incorporating instrumented tactile feedback and its impact on policy performance and adaptability in complex manipulation tasks. Our results show that pre-training with human demonstrations improves the efficiency of downstream imitation learning, achieving better performance with fewer robot demonstrations and increased robustness across considered tasks.

\subsection{Task and Demonstrations} We consider the task of grasping an arbitrarily oriented object on a table using a gripper equipped with tactile sensing. To achieve this, we collect demonstrations from both humans and the robot, as shown in Fig \ref{fig:demosetup}. The human dataset includes five objects, with 1,000 human demonstrations per object. Since robot demonstrations require teleoperation, which is more challenging and time-consuming, we collect them for only 100 demos on a single object. The objects used for pre-training with human demos, fine-tuning, and generalization are all shown in Fig \ref{fig:demo-objects}.

\subsection{Task Performance and Generalization with Few Robot Demonstrations}

In this section, we evaluate the improvement in demonstration complexity achieved through pre-training with our method. We compare our approach against baseline methods that use different pre-trained encoders, including alternative strategies for pre-training with human demonstrations. For these experiments, tactile input is not used, allowing us to focus solely on the impact of pre-training on performance. We evaluate the following methods:

\subsubsection{Diffusion Policy (DP)}
This baseline is a diffusion policy using a ResNet-18 encoder to encode each view in history of observations. The ResNet is initialized with ImageNet pre-trained weights, serving as a baseline without task-specific pre-training. Note that weights from ImageNet pre-training are used exclusively for the this baseline and not in any other baselines. This baseline does not use any human demonstrations.

\subsubsection{Diffusion Policy with ViTacT Encoder Pretrained on Human Demos Using Time-Contrastive Loss (DP + TCL)} In this baseline, we consider the Time Constrastive Loss as proposed by R3M \cite{Nair2022-ss} baseline. Temporally relevant features are important for manipulation skills. Therefore, we consider a common temporal contrastive learning technique for learning state representations. We train our ViTacT encoder using our human demonstration dataset with a temporal modeling approach based on Time Contrastive Loss.  Latents for temporally closer images from a video demonstration have higher similarity compared to images that are temporally distant or come from different videos using contrastive learning. Specifically, a batch of frame sequences is sampled, consisting of frames $\mathbf{o}_i$, $\mathbf{o}_j$ (where $j > i$), and $\mathbf{o}_k$ (where $k > j$) is sampled and the InfoNCE loss is minimized.
\[
\mathcal{L}_{\text{TCL}} = - \sum_{b \in \mathcal{B}} \log \frac{e^{d(x_i^b, x_j^b)}}{e^{d(x_i^b, x_j^b)} + e^{d(x_i^b, x_k^b)} + e^{d(x_i^b, x_{i}^{\neq b})}},
\] 
where $x$ is the latent state and $x_{i}^{\neq b}$ is a latent state from a
$\mathbf{o}_i^{\neq b}$ - observation  from a different demo in the batch. $d$ denotes a measure of similarity, where we specify as the negative $l_2$ distance.

\subsubsection{Diffusion Policy with JIF Pretraining (DP + JIF)} this is our method.

\begin{figure}[t]
    \centering
    \includegraphics[width=\linewidth]{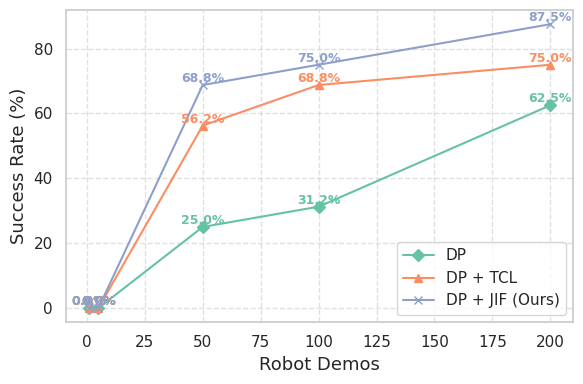}
    \caption{Grasping success rate of our method against the baselines. While both pre-training approaches improve demonstration complexity, with JIF pre-training we achieve the highest success rate.}
    \label{fig:results-1}
\end{figure}

The first metric we consider for all methods above is task success as a function of the number of robot demonstrations used in training. Higher success rate with fewer robot demonstrations implies that a method is better able to extract value from human pre-training. We evaluate all methods on a number of robot demonstrations varying between 1 and 200. The success rate of the resulting imitation policies for each method is measured over 15 trials. A detailed performance comparison is presented in Fig.~\ref{fig:results-1}. 

We note that our method (DP + JIF) achieves more than twice the success rate compared to the DP baseline without human pre-training. Additionally, DP + JIF demonstrates superior sample efficiency, achieving a higher success rate compared to the DP + TCL baseline when using a similar number of robot demonstrations. These results underscore the benefits of incorporating human pre-training via JIF in enhancing both success rates and data efficiency.

\begin{table}[t]
    \centering
    \caption{Generalization Performance. We compare the generalization performance of our method DP + JIF fine-tuned with 50 robot demonstrations on grasping objects out-of-distribution w.r.t fine-tuning robot demonstrations.}
    \renewcommand{\arraystretch}{1.2} 
    \setlength{\tabcolsep}{4pt} 
    \resizebox{\columnwidth}{!}{ 
    \begin{tabular}{lccccc}
        \toprule
        \textbf{Object} & \textbf{Eggplant} & \textbf{Banana} & \textbf{Lego} & \textbf{Screwdriver} & \textbf{Hammer} \\
        \midrule
        \textbf{Method} & {\color{ForestGreen}In-distribution} & \multicolumn{4}{c}{\color{BrickRed} Out-of-distribution $\rightarrow$} \\
        \midrule
        DP & 25.0\% & 37.5\% & 43.8\% & 12.5\% & 31.3\% \\
        DP + TCL & 56.3\% & 62.5\% & 62.5\% & 50.0\% & \textbf{56.3\%} \\
        \textbf{DP + JIF} (Ours) & \textbf{68.8\%} & \textbf{68.8\%} & \textbf{68.8\%} & \textbf{50.0\%} & 43.8\% \\
        \bottomrule
    \end{tabular}
    }
    \label{tab:generalization_performance}
\end{table}

We then assess the generalization performance of our approach. Specifically, we evaluate the success rate on objects that were not included in the robot demonstrations, as detailed in Table I. For each training method, we use the model trained on 50 demonstrations. 

Our data shows that the model can generalize even without human data pre-training. However, the TCL and JIF models, both pre-trained with human data, achieve higher success rates, with our JIF model outperforming the others on most objects.

\subsection{Robustness with Visuo-tactile Human Demonstrations}

\begin{figure}[t]
    \centering
    \includegraphics[width=0.98\linewidth]{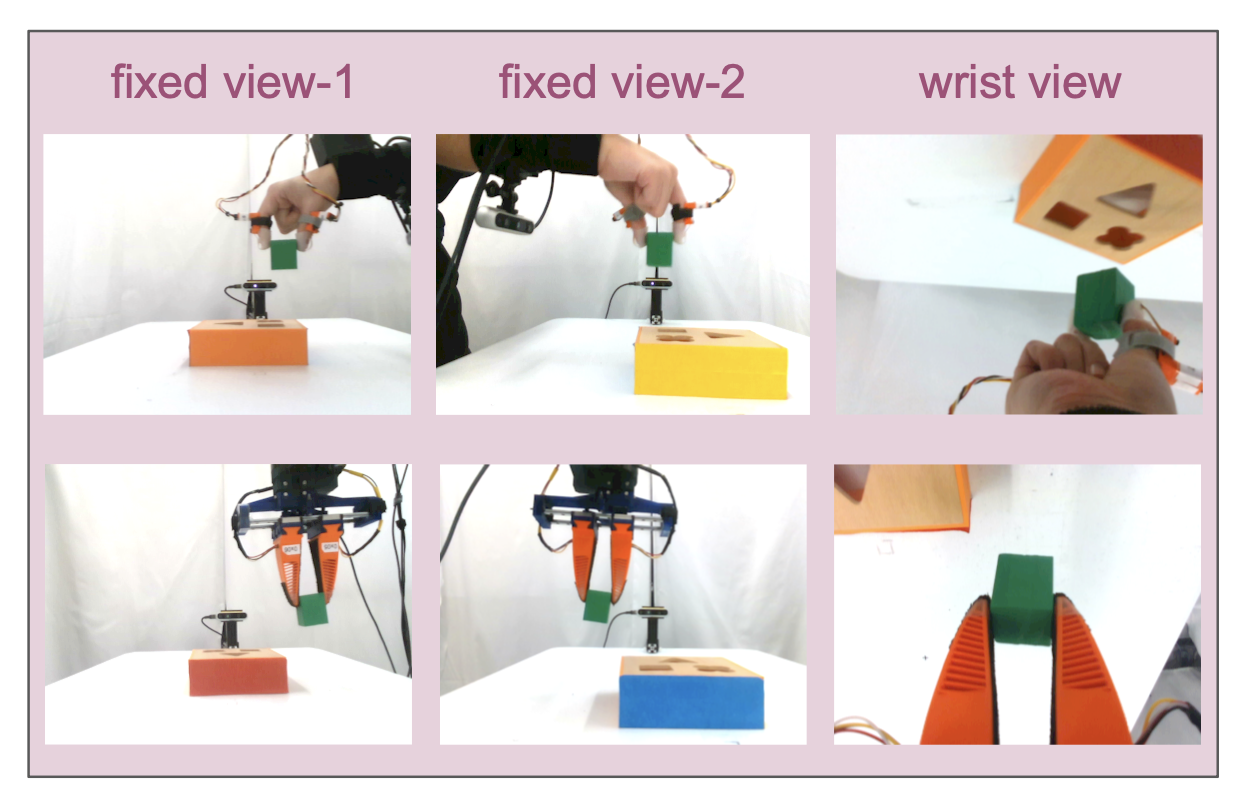}
     \caption{The importance of instrumented human demonstrations with tactile sensing is evaluated using a challenging peg-in-hole insertion task. The figure illustrates both human and robot demonstrations during the placement of a cube into a square hole.}
    \label{fig:insertion_task}
\end{figure}


To evaluate the impact of instrumented human demonstrations, we incorporate tactile readings as additional inputs for state representation learning and evaluate with a peg-in-insertion task. As before we collect a 1,000 human demonstrations with 100 robot demonstrations, as shown in Fig \ref{fig:insertion_task}. We compare the following tactile sensing conditions:

\subsubsection{No Tactile Sensing}
Both human demonstrations and the robot policy rely solely on visual data without incorporating tactile inputs.


\subsubsection{Human and Robot Tactile Sensing}
Both human demonstrations and the robot policy incorporate tactile sensing alongside visual data, providing a richer representation of the manipulation process.

\begin{figure}[t]
    \centering
    \includegraphics[width=\linewidth]{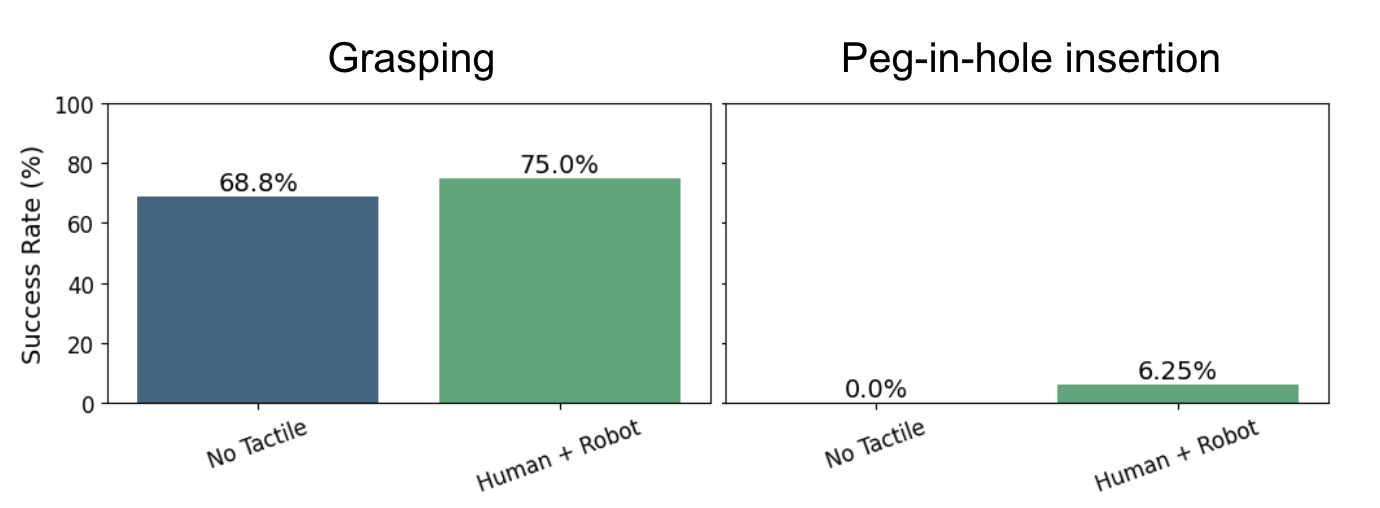}
     \caption{The results underscore the importance of instrumented human demonstrations with tactile sensing, as evidenced by a success rate comparison across different tactile sensing configurations. The left plot shows notable improvements in grasping success rates, while the right plot highlights the successful execution of the challenging peg-in-hole insertion task, albeit with a lower success rate. Incorporating tactile feedback, particularly in both human and robot demonstrations, enhances performance and achieves higher success rates with fewer robot demonstrations.}
    \label{fig:results-tactile-ablation}
\end{figure}

The results of this comparison are summarized in Fig.~\ref{fig:results-tactile-ablation}. As shown, pre-training with tactile sensing leads to a notable improvement in success rates for grasping, demonstrating the value of incorporating tactile feedback in both human and robot demonstrations. More importantly, this approach enables the successful execution of the challenging task of peg-in-hole insertion, albeit with a low success rate, emphasizing the role of instrumented human demonstrations in learning complex, contact-rich tasks.

\begin{figure}[t]
    \centering
    \includegraphics[trim=28mm 97mm 7mm 0mm,clip=true,width=0.49\columnwidth]
    {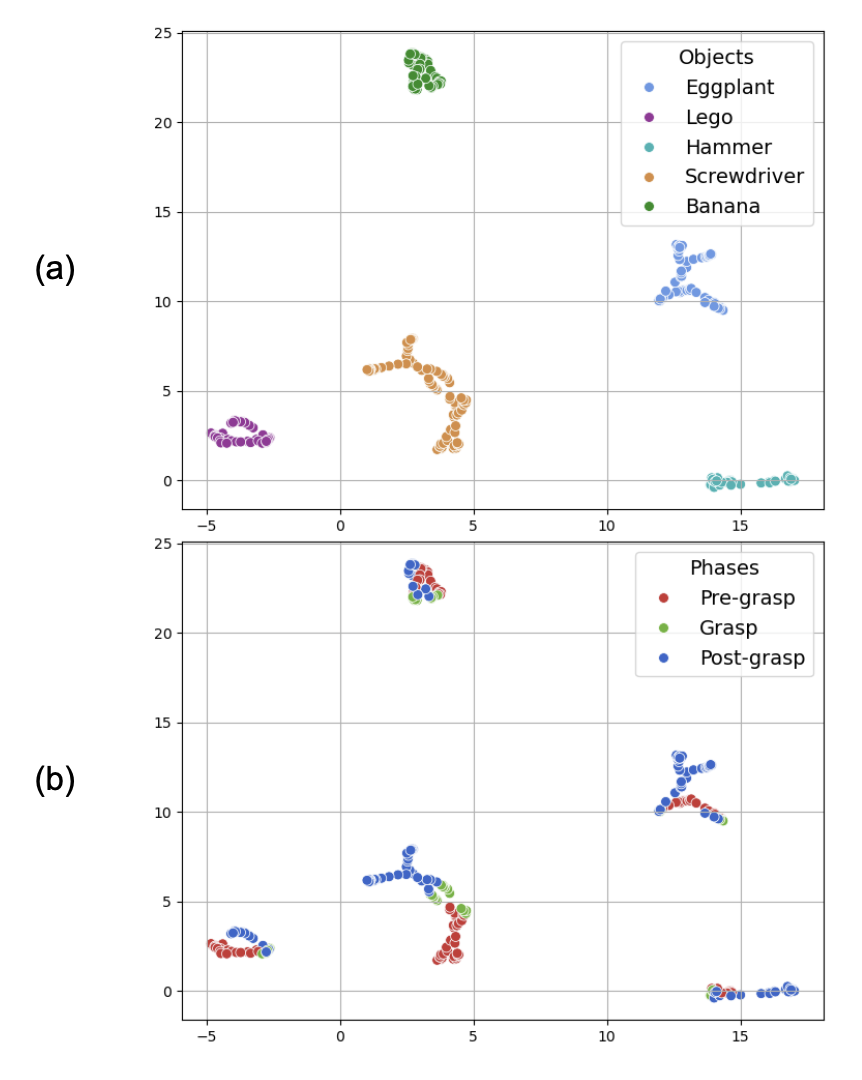}
    \includegraphics[trim=28mm 7mm 7mm 94mm,clip=true,width=0.49\columnwidth]
    {images/umap/umap_combined_vertical.png}    
    \caption{UMAP visualization of the learned representations. Left: The representations exhibit clear separability among the five different objects, indicating that the model effectively captures object-specific features. Right: The latent space shows meaningful structure with respect to task phases, distinctly separating the pre-grasp, grasp, and post-grasp phases. This suggests that the learned representations capture manipulation-specific information.}
    \label{fig:umap}
\end{figure}

\subsection{Visualizing Learned Latent States}

 We applied Uniform Manifold Approximation and Projection (UMAP) \cite{McInnes2018-cu} to visualize the embedding space. We utilized two demonstrations per object to assess whether distinct object categories naturally emerge within the latent space. The UMAP visualization (Fig \ref{fig:umap}) revealed clear separability among the five different objects, indicating that the learned representations effectively capture object-specific features. Furthermore, the latent space exhibited meaningful structure with respect to task phases, as it distinctly separated the pre-grasp, grasp, and post-grasp phases. 
 

\subsection{Limitations}

While this study provides valuable insights into leveraging human demonstrations for pre-training in imitation learning, several limitations should be considered. First, the study's scale is limited to a reduced number of tasks. Furthermore, despite variations in inertial properties, the target objects used for grasping share similar aspect ratios, potentially limiting the diversity of learned representations. Additionally, the focus on grasping—a relatively simple task—may not fully reflect the complexities of real-world manipulation scenarios.

Methodologically, the reliance on human demonstrations does present its own challenges in terms scalability due to the need for instrumentation (cameras, wearable tactile sensors) which could limit its applicability "in the wild". While one possible direction for the field aims to completely eliminate the need for any robot demonstrations, our approach still depends on robot demonstrations for fine-tuning, which can be impractical in scenarios where robotic data collection is infeasible or costly. Nevertheless, we believe that the ability to extract value from human demonstrations, complementing a much smaller number of robot demonstrations, can offer a significant advantage towards scalability in deployment.

\section{Discussion and Future Work}

In this work, we introduced a novel pre-training framework that leverages multi-modal human demonstrations to acquire manipulation-centric representations for imitation learning. Our approach provides several key advantages, including improved fine-tuning efficiency, enhanced generalization, and greater computational efficiency by avoiding costly action annotations and high-dimensional reconstructions. These findings position human demonstration-based pre-training as a promising solution to critical challenges in imitation learning, particularly in addressing the pervasive issue of data scarcity.

A key contribution of our work is the use of instrumented multi-modal human demonstrations, particularly with tactile sensing, which significantly improves performance. This highlights the potential of richer sensory inputs, such as touch, in advancing robotic manipulation capabilities and closing the gap between human and robot skill acquisition.

We aim to extend our framework by incorporating latent actions alongside latent states during fine-tuning to further enhance policy learning efficiency. Additionally, we plan to explore the integration of instrumented human demonstration pre-training within offline reinforcement learning to better leverage human priors for data-efficient policy optimization.

We believe that multi-modal human demonstrations, especially those enriched with tactile and proprioceptive feedback, will continue to play a crucial role in developing robust robotic systems. As many valuable representations of the physical world are inherently grounded in sensory interaction, learning through touch and other modalities will be pivotal for advancing robotic dexterity. Future research focus on scaling our approach to diverse, real-world tasks and addressing the challenges posed by complex, high-dimensional manipulation scenarios seems promising.